\documentclass{article}

\usepackage{arxiv}

\usepackage[utf8]{inputenc} 
\usepackage[T1]{fontenc}    
\usepackage{hyperref}       
\usepackage{url}            
\usepackage{booktabs}       
\usepackage{amsfonts}       
\usepackage{nicefrac}       
\usepackage{microtype}      
\usepackage{graphicx}
\usepackage{natbib}
\usepackage{doi}
\usepackage{amsmath}
\usepackage{titlesec}
\usepackage{tabularx}
\usepackage{adjustbox} 
\usepackage{tikz}
\usepackage{xr}
\usepackage{multirow}
\makeatletter
\renewcommand\subsection{\@startsection{subsection}{2}{\z@}%
  {-3.25ex\@plus -1ex \@minus -.2ex}%
  {1.5ex \@plus .2ex}%
  {\normalfont\normalsize}}
\makeatother

\makeatletter
\renewcommand\subsubsection{\@startsection{subsubsection}{2}{\z@}%
  {-3.25ex\@plus -1ex \@minus -.2ex}%
  {1.5ex \@plus .2ex}%
  {\normalfont\normalsize}}
\makeatother

\titleformat{\section}{\normalfont\Large\bfseries}{\thesection}{1em}{}
\titleformat{\subsubsection}[runin]{\normalfont\normalsize\bfseries}{\thesubsubsection}{1em}{}[:]
\titlespacing*{\subsubsection}{0pt}{\baselineskip}{0.5\baselineskip}

\title{ADEP: A Novel Approach Based on Discriminator-Enhanced Encoder-Decoder Architecture for Accurate Prediction of Adverse Effects in Polypharmacy}

\author{ 
Katayoun Kobraei \\
Department of Computer and Data Sciences\\
Faculty of Mathematical Sciences, Shahid Beheshti University\\
Tehran, Iran \\
\texttt{} \\
\And
Mehrdad Baradaran \\
Department of Computer and Data Sciences\\
Faculty of Mathematical Sciences, Shahid Beheshti University\\
Tehran, Iran \\
\texttt{} \\
\And
Seyed Mohsen Sadeghi \\
Department of Computer and Data Sciences\\
Faculty of Mathematical Sciences, Shahid Beheshti University\\
Tehran, Iran \\
\texttt{} \\
\And
\href{https://orcid.org/0000-0001-8166-3815}{\includegraphics[scale=0.06]{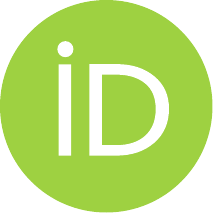}\hspace{1mm}Raziyeh Masumshah} \\
Department of Computer and Data Sciences\\
Faculty of Mathematical Sciences, Shahid Beheshti University\\
Tehran, Iran \\
\texttt{} \\
\And
\href{https://orcid.org/0000-0002-8913-3904}{\includegraphics[scale=0.06]{orcid.pdf}\hspace{1mm}Changiz Eslahchi}\thanks{Corresponding author} \\
Department of Computer and Data Sciences\\
Faculty of Mathematical Sciences, Shahid Beheshti University\\
Tehran, Iran \\
School of Biological Sciences\\
Institute for Research in Fundamental Sciences (IPM)\\
Tehran, Iran \\
\texttt{ch-eslahchi@sbu.ac.ir} \\
}

\hypersetup{
pdftitle={ADEP: A Novel Approach Based on Discriminator-Enhanced Encoder-Decoder Architecture for Accurate Prediction of Adverse Effects in Polypharmacy},
pdfauthor={Katayoun Kobraei, Mehrdad Baradaran, Seyed Mohsen Sadeghi, Raziyeh Masumshah, Changiz Eslahchi},
pdfkeywords={Polypharmacy, Adverse Effects, Encoder-Decoder Architecture, Discriminator, Machine Learning},
}

\begin{document}
\maketitle

\begin{abstract}
Motivation: Unanticipated drug-drug interactions (DDIs) pose a significant risk of severe bodily harm, emphasizing the importance of predicting adverse effects in polypharmacy. Recent advancements in computational methods have aimed to tackle this critical challenge.\\
Methods: This paper introduces ADEP, a novel approach that integrates a discriminator and an encoder-decoder model to address data sparsity and improve feature extraction accuracy. ADEP utilizes a three-part model including a multiple classification method to predict adverse effects in polypharmacy.\\
Results: Performance evaluation on three benchmark datasets demonstrates that ADEP outperforms several well-known methods, including GGI-DDI, SSF-DDI, LSFC, DPSP, GNN-DDI, MSTE, MDF-SA-DDI, NNPS, DDIMDL, Random Forest (RF), K-Nearest-Neighbor (KNN), Logistic Regression (LR), and Decision Tree (DT). Comprehensive analysis of ADEP's performance on these datasets highlights key metrics such as Accuracy (ACC), Area Under the Receiver Operating Characteristic Curve (AUROC), Area Under the Precision-Recall Curve (AUPRC), F-score, Recall, Precision, False Negatives (FN), and False Positives (FP). The results demonstrate that ADEP achieves more accurate prediction of adverse effects associated with polypharmacy. Furthermore, a case study conducted with real-world patient data illustrates the practical application of ADEP in identifying potential drug-drug interactions and preventing adverse effects.\\
Conclusions: ADEP represents a significant advancement in predicting adverse effects of polypharmacy, offering improved accuracy and reliability. Its innovative architecture and alternative classification methods contribute to better representation and feature extraction from sparse medical data, thereby enhancing medication safety and patient outcomes.
Availability: The source code and datasets are available at \href{https://github.com/m0hssn/ADEP}{https://github.com/m0hssn/ADEP}.
\end{abstract}


\section{Introduction}
In clinical practice, there has been a significant rise in the adoption of multi-drug combination therapy, commonly referred to as polypharmacy. This therapeutic approach is increasingly favored for its acknowledged capacity to enhance treatment effectiveness and expand the range of available treatment options \citep{Su2022, Sutherland2015, Guillot2020}. With the rapid growth in the number of approved drugs, it has become commonplace to prescribe multiple medications simultaneously for patient treatment. Polypharmacy can lead to unwanted drug-drug interactions (DDIs), wherein the pharmacological effect of one drug is altered by another. A reported 35.8\% of elderly Americans aged 61 to 80 concurrently used at least five prescription drugs, with approximately 15.1\% of them facing potential risks of major drug-drug interactions due to medication combinations \citep{Qiu2021, Chen2016a, Zitnik2018, Bumgardner2021, Lin2022a}. Adverse drug reaction events in the United States alone incur an annual expenditure exceeding \$10 billion, with over 30\% of these costs attributed to DDIs. DDIs are the most common reason for patients to seek emergency care and can lead to Adverse Drug Reactions (ADRs), including fatal outcomes, making them a critical public health concern \citep{Shtar2019, Lin2020b}. As a result, promptly identifying drug-drug interactions (DDIs) is crucial for mitigating potentially life-threatening risks and enhancing both drug development efforts and clinical treatment protocols. Due to the vast number of potential drug combinations, conventional approaches like in vitro studies and clinical trials, known for their time-consuming, inefficient, and costly nature, encounter obstacles in accurately detecting DDIs \citep{Whitebread2005, Rudrapal2020}. Computational approaches have emerged as valuable alternatives, offering efficiency and cost-effective advantages in the identification of potential DDIs \citep{Tatonetti2012, Yu2022a, Deng2022, Chen2021b}. These methods, predominantly characterized by their learning objectives, encompass four primary categories: feature-based, similarity-based, network-based, and matrix decomposition-based approaches \citep{Lin2023c, Yu2024b}. By leveraging various drug characteristics such as chemical structure, side effects, and target information, feature-based methods aim to predict potential DDIs \citep{Huang2020, Zhang2015a, Zhang2017b}. For instance, NNPS \citep{Masumshah2021a} incorporates monoside effects and drug-protein interaction data into drug representations, amalgamating them for DDI prediction. Lin et al. introduced the MDF-SA-DDI method, which relies on multi-source drug fusion, multi-source feature fusion, and a transformer self-attention mechanism to facilitate latent feature fusion \citep{Lin2022d}. The LSFC, proposed by Zhou et al., is a multichannel feature fusion method that leverages local substructure features of drugs and their complements. These local substructure features are extracted from each drug, interacted with those of another drug, and subsequently integrated with the global features of both drugs to facilitate DDI prediction \citep{Zhou2023}. Similarity-based methods operate under the assumption that drugs with similar characteristics are likely to share common interactions. These approaches compute similarity scores between drugs and represent them as similarity vectors. Prediction models then utilize these vectors to identify potential DDIs \citep{Deng2020b, Gottlieb2012}. The DDIMDL method constructs sub-models utilizing a wide range of drug features and integrates them into a deep neural network (DNN) framework. These sub-models are subsequently merged and inputted into fully connected neural networks, enabling comprehensive prediction \citep{Deng2020b}. DPSP offers an approach by integrating multiple features of drug information, such as mono side effects, target proteins, enzymes, chemical substructures, and pathways. This comprehensive assessment enhances the understanding of drug similarity and aids in predicting drug-drug DDIs more effectively \citep{Masumshah2023b}. Network-based methods, conversely, often establish a knowledge graph of DDIs and derive drug embeddings by analyzing their local or global topological properties within the graph. This approach enables the inference of potential DDIs based on network connectivity \citep{Wang2020}. Yao et al. introduced a novel knowledge graph embedding technique called MSTE, which transforms the side effects prediction task into a link prediction problem \citep{Yao2022}. Similarly, the GNN-DDI method developed a drug embedding vector and employed a graph neural network to predict side effects associated with polypharmacy \citep{Al-Rabeah2022}. The SSF-DDI model integrates drug sequence features and structural characteristics from the drug molecule graph. This incorporation enhances the information available for DDI prediction, resulting in a more comprehensive and accurate representation of drug molecules \citep{Zhu2024}. In the GGI-DDI method, drugs are aggregated into a series of larger granules that encapsulate the key substructures or functional groups of drugs. DDIs are then detected by analyzing interactions among these granules, mirroring human cognitive patterns more closely \citep{Yu2024b}. Matrix decomposition-based methods involve factorizing the adjacency matrix of DDIs into multiple factor matrices. These factor matrices are then used to reconstruct the adjacency matrix, enabling the identification of potential DDIs \citep{Narita2012}. While existing models have demonstrated promising outcomes in predicting DDIs, a number of challenging issues endure:
\begin{itemize}
\item Interpretability: Many DDIs prediction models, especially those leveraging deep learning techniques, encounter practical constraints owing to their limited interpretability. It remains essential to elucidate the decision-making mechanisms of these models in a manner that resonates with human comprehension.
\item Interaction Extrapolation: Real-world applications frequently necessitate models to extrapolate interactions not solely between known drugs and new drugs but also among entirely novel drugs. Regrettably, numerous methods, particularly those grounded in network-based and matrix decomposition-based approaches, encounter difficulties in effectively scrutinizing such interactions.
\end{itemize}
Hence, there is a pressing need to devise a novel methodology capable of overcoming these constraints and offering a more comprehensive and dependable approach to predicting drug-drug interactions. In this study, we present a novel and comprehensive approach to addressing the challenges associated with polypharmacy side effects. Our proposed method, known as an Approach based on Discriminator-enhanced Encoder-decoder architecture for accurate prediction of adverse effects in Polypharmacy (ADEP), aims to overcome the limitations highlighted above. ADEP integrates an autoencoder with a classifier and a discriminator, thereby offering a multifaceted solution for robust training.

ADEP is designed to fulfill multiple objectives, including feature extraction, classification, and adversarial training. To achieve this, ADEP adopts a structured three-step process during training: feature extraction, adversarial training and classification. By leveraging these components in tandem, ADEP provides a versatile and powerful framework for effectively predicting drug interactions and mitigating the risks associated with polypharmacy. The following section provides an overview of the datasets utilized in this study, along with a detailed description of the ADEP method. Section 3 presents a comparative analysis of the results obtained from ADEP and other existing methods, employing six distinct criteria for evaluation. Finally, Section 4 offers the conclusion drawn from our findings, as well as outlines potential avenues for future research.

\section{Material and Method}
In this section, we provide a comprehensive overview of the method, detailing the input format and all its steps. The overall framework of ADEP is depicted in Figure \ref{fig1}.
\subsection{Datasets}

Our proposed model leverages three benchmark datasets, DS1, DS2, and DS3, each containing distinct DDI adverse events. These datasets collectively enable a comprehensive evaluation of our model's performance across various types of drug interactions and associated adverse events. DS1 comprises information on 572 drugs with a total of 37,264 known DDIs, each linked to 65 types of adverse events. The dataset's features include comprehensive drug-related information, such as mono side effects, targets, enzymes, chemical substructures, and pathways. These features were curated from reputable databases, including DrugBank, KEGG, and Pubchem \citep{Knox2024, Kanehisa2000, Kim2023}. Furthermore, mono side effect data were integrated from the OFFSIDES and SIDER databases \citep{Kuhn2016, Tatonetti2012}. DS2 involves 1258 drugs and documents 161,770 known DDIs associated with 100 types of adverse events. For this dataset, drug features, including targets, enzymes, and chemical substructures, were exclusively collected from the DrugBank database. It is important to note that the DS1 and DS2 datasets are imbalanced, with one event comprising 10,000 drug pairs while another event contains only 100 drug pairs. DS3 introduces a unique perspective by including information on 645 drugs and 63,473 drug interactions. It's worth noting that each drug pair in DS3 may exhibit multiple adverse effects. In our analysis, we focused on the side effects with the highest frequency for drug combinations with multiple adverse effects. Specifically, we considered 185 side effects based on the paper \cite{Masumshah2023b}, ensuring a comprehensive evaluation of potential adverse outcomes associated with each drug combination in DS3. These datasets afford a thorough assessment of the effectiveness of our proposed ADEP model. The detailed characteristics of each dataset, including adverse events and drug features, are concisely outlined in Table \ref{tab1}.

\begin{table*}[ht]
\caption{Details of benchmark datasets.}
\label{tab:benchmark_details}
\resizebox{16cm}{!}{
\begin{tabular*}{\linewidth}{@{\extracolsep{\fill}}>{\centering}p{2.7cm}>{\centering}p{4.5cm}>{\centering}p{4.5cm}l@{}}
\toprule
\textbf{Benchmark name} & \textbf{Benchmark details} & \textbf{Benchmark features} & ~~~~~~~~\textbf{Reference} \\
\midrule
DS1 & \begin{tabular}[t]{@{}l@{}}~~~~~No. drugs=572 \\ No. interactions=37264 \\ ~~~~~No. events=65\end{tabular} & \begin{tabular}[t]{@{}c@{}}\small No. side effects=9991 \\ No. targets=1162 \\ No. enzymes=202 \\ No. chemical substructure=881 \\ No. pathways=957\end{tabular} & \begin{tabular}[t]{@{}c@{}}\footnotesize\citep{Deng2020b} \\ \footnotesize\citep{Al-Rabeah2022}\end{tabular} \\
\addlinespace[1.5ex] 
DS2 & \begin{tabular}[t]{@{}l@{}}~~~~No. drugs=1258 \\ No. interactions=161770 \\ ~~~~~No. events=100\end{tabular} & \begin{tabular}[t]{@{}c@{}}No. targets=1651 \\ No. enzymes=316 \\ No. chemical substructure=2040\end{tabular} & ~~~~\footnotesize\citep{Lin2022d} \\
\addlinespace[1.5ex] 
DS3 & \begin{tabular}[t]{@{}l@{}}~~~~~No. drugs=645 \\ No. interactions=63473 \\ ~~~~~No. events=185\end{tabular} & \begin{tabular}[t]{@{}c@{}}No. side effects=10184 \\ No. targets=8934\end{tabular} & ~~~~\footnotesize\citep{Masumshah2021a} \\
\bottomrule
\end{tabular*}}
\label{tab1}
\end{table*}

\subsection{Feature Vector Integration}
In the initial step, for each drug, we generate a vector by concatenating various features. Then, for the two specified drugs, we concatenate their respective vectors. We present the drug interactions to the model as pairs of drug information, both in the order of 'drugA' and 'drugB', as well as in the reverse order of 'drugB' and 'drugA'. For each drug pair, we create two feature vectors as the input of the model, with sizes of 26,386 in DS1, 8,014 in DS2, and 38,236 in DS3.

\subsection{ADEP Method}
The ADEP model integrates an autoencoder with a classifier and a discriminator for comprehensive training. Developed to serve the purposes of feature extraction, classification, and adversarial training, it follows a structured three-step process during these phases.
In the feature extraction phase, the encoder is tasked with encoding the input and generating the latent space such that it fulfills several objectives. Firstly, it should facilitate the reconstruction of the input data with minimal loss through the decoder. Secondly, the discriminator should leverage the appropriate latent space to educate the model in discerning disparities among various classes and distinguishing data points within the space. Lastly, the latent space should be structured in a manner conducive to effective utilization for classification purposes. This decoding process is pivotal for effective feature extraction, emphasizing the extraction of deep features from the latent space. This strategy emphasis on feature extraction over raw data preservation enhances the model's capacity for accurate class prediction.
The model accommodates imbalanced data scenarios. To counteract this imbalance, a discriminator is introduced as part of the model architecture. In addition to its conventional roles in the preceding training phases, the discriminator plays a crucial role in representing the latent space in a manner that spatially distinguishes each class from others. The outcome of this spatial differentiation is that the representation of each class is confined to a distinct region, streamlining the classifier's task and contributing to the overall efficacy of ADEP in handling imbalanced datasets.
In the following we explain each component in detail.

\begin{figure*}[htbp]
    \centering
    \begin{tikzpicture}
        \node[anchor=south west,inner sep=0] (image) at (0,0) {\includegraphics[width=1\textwidth]{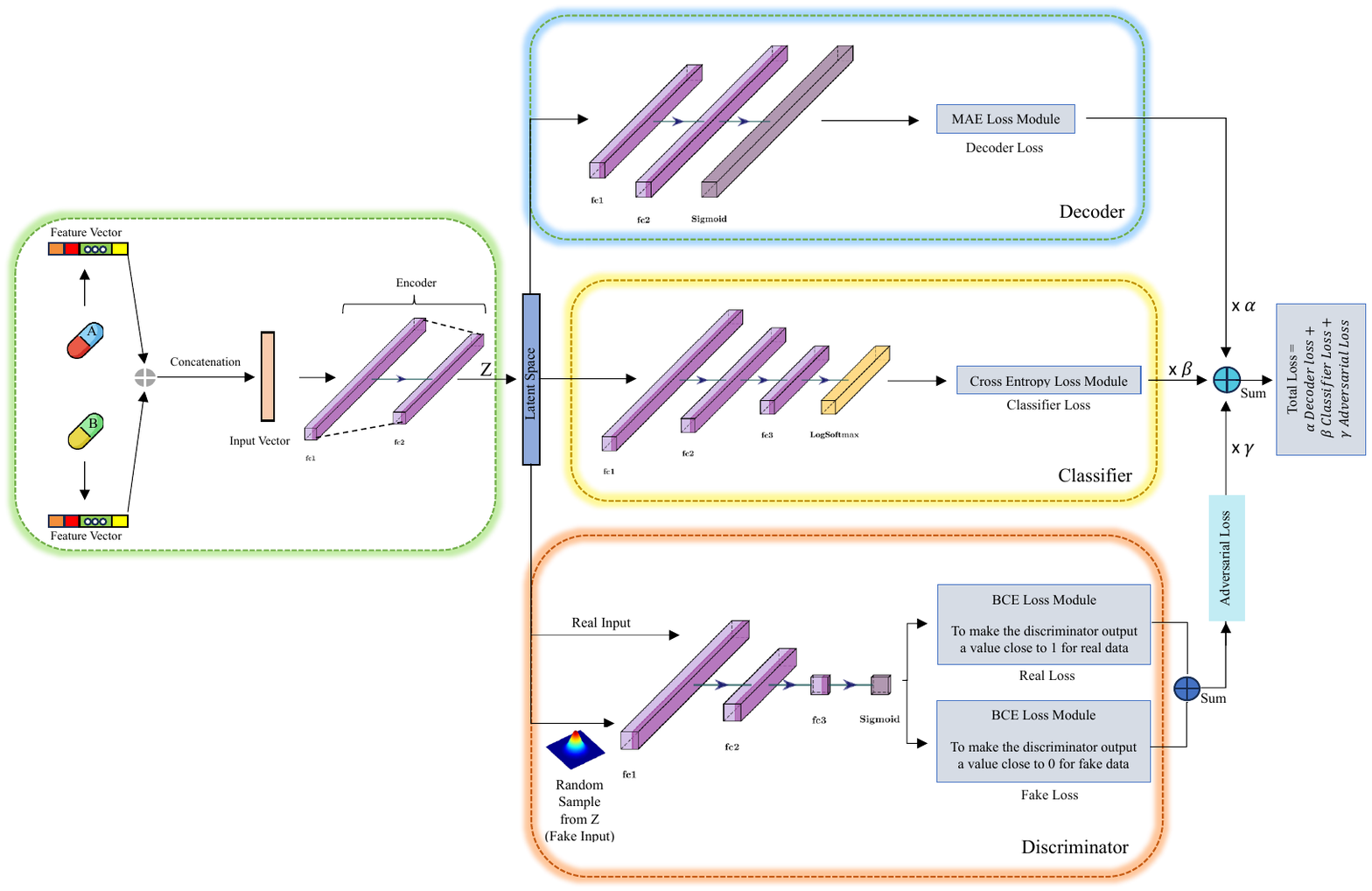}}; 
        \begin{scope}[x={(image.south east)},y={(image.north west)}]
            \node[font=\scriptsize] at (0.027,0.70) {(a)}; 
            \node[font=\scriptsize] at (0.40,0.89) {(b)}; 
            \node[font=\scriptsize] at (0.403,0.64) {(c)};
            \node[font=\scriptsize] at (0.405,0.4) {(d)};
            \node[font=\scriptsize] at (0.96,0.64) {(e)};
        \end{scope}
    \end{tikzpicture}
    \caption{The workflow of ADEP. a) Integration of drug feature vectors into a unified input vector through an encoder, facilitating comprehensive data representation for subsequent use by decoder, classifier, and discriminator components. b) Leveraging latent space for classification tasks and accurate input vector recreation by decoder. c) Utilization of latent space by classifier model for classification and prediction tasks. d) Generating samples from the latent distribution, training the model to distinguish between classes, and addressing data imbalance by identifying fake data closely resembling the original by the discriminator. e) Aggregate loss calculation: combining auxiliary losses from all three parts with coefficients to obtain final loss for backpropagation.}
    \label{fig1}
\end{figure*}

\subsubsection*{$Module 1: Feature Extraction$}

In this module, the encoder and decoder are responsible for feature extraction. Another outcome of this process is dimensionality reduction. The latent representation has fewer dimensions than the original data, making it a more compact and efficient representation.

\paragraph{Encoder}

The encoder takes the input and reduces it to a more compact and meaningful lower-dimensional representation (Figure \ref{fig1}a). This representation serves as a distilled form of the original data, capturing essential patterns and features while reducing its dimensionality. The representation of the encoder is known as the latent representation or feature vector. The encoder comprises two fully connected (linear) layers. The initial layer processes input features, yielding an output with 4,096 units. Subsequently, the second layer compresses the 4,096 units from the preceding layer into a lower-dimensional representation containing 2,048 units. After each fully connected layer, batch normalization is applied to enhance training stability and accelerate convergence. Rectified Linear Unit (ReLU) activation functions follow each fully connected layer to introduce non-linearity and capture complex data patterns. Dropout is used as a regularization technique, with a rate of 0.3 applied after the first layer and a rate of 0.2 applied after the second layer.

\paragraph{Decoder}

The decoder's role is to reconstruct the input data from the lower-dimensional representation based on the information contained in the latent representation (Figure \ref{fig1}b). The reconstruction loss, typically measured as the difference between the original data and the reconstructed data, is minimized during training. By minimizing this loss, the autoencoder learns to capture essential features and patterns in the data, as the reconstruction process encourages the network to focus on the most informative aspects. The decoder consists of two fully connected layers, similar to the encoder. The first layer takes the 2,048 units from the encoder's second layer and maps them to a 4,096 units representation. The second layer takes this 4,096 units representation and expands it back to the original input dimension. Like the encoder, batch normalization is applied after each fully connected layer for training stability. ReLU activation functions follow each fully connected layer to introduce non-linearity. Dropout with a rate of 0.3 is applied after the first layer in the decoder. Batch normalization and ReLU activation functions are used, and a final Sigmoid activation function ensures the output is within the range [0, 1]. The decoder utilizes a Mean Absolute Error (MAE) loss module. The MAE loss measures the absolute difference between the original data and the reconstructed data. This loss function guides the training process by penalizing the model based on the average magnitude of the errors in the reconstruction, contributing to the overall learning objective of the autoencoder. The MAE loss, also interchangeably referred to as the autoencoder loss in this study, is calculated using the following formula:

\begin{equation}
\label{eq:mae}
\text{MAE} = \frac{1}{N} \sum_{i=1}^{N} \left| y_i - \hat{y}_i \right|
\end{equation}
where $N$ is the number of samples in the dataset, $y_i$ is the true value for sample $i$, and $\hat{y}_i$ is the predicted value for sample $i$.

\subsubsection*{$Module 2: Classifier$}

At the outset, the classifier operates by taking the encoded representation as input and generating class predictions (Figure \ref{fig1}c). It consists of three fully connected layers. The first layer takes the 2,048 dimensional encoded representation as input and reduces it to 512 dimensions. The second layer further reduces it to 256 dimensions. Finally, the output layer produces 65 output classes for DS1, 100 for DS2, and 185 for DS3. Batch normalization is applied after the first and second fully connected layers in the classifier. This helps stabilize training and improve convergence. ReLU activation functions are applied after each fully connected layer. ReLU introduces non-linearity, allowing the model to capture complex data patterns. A Dropout layer with a rate of 0.2 is applied after the first fully connected layer to prevent overfitting. A LogSoftmax activation function is applied to the output of the final layer. This function converts the raw scores into a probability distribution over the classes, making it suitable for multiclass classification tasks. Additionally, the classifier employs cross-entropy loss, a commonly used loss function for multiclass classification problems. Cross-entropy loss measures the difference between the predicted probability distribution and the true distribution of the classes, guiding the model to minimize this difference during training. The cross-entropy loss, often termed as the classifier loss in this study, is calculated using the following formula:

\begin{equation}
\label{eq:cross_entropy}
\text{Cross-Entropy Loss} = -\frac{1}{N} \sum_{i=1}^{N} \sum_{j=1}^{C} y_{ij} \cdot \log(p_{ij})
\end{equation}

where $N$ is the number of samples in the dataset, $C$ is the number of classes, $y_{ij}$ is a binary indicator (0 or 1) if class $j$ is the correct classification for sample $i$, and $p_{ij}$ is the predicted probability that sample $i$ belongs to class $j$ according to the model.

\subsubsection*{$Module 3: Discriminator$}

The discriminator is employed for adversarial training. Adversarial training is a technique commonly used in generative adversarial networks (GANs) and, in this case, in ADEP \citep{Creswell2018, Makhzani2015}. The primary goal is to make the encoded representations produced by the encoder (the real representations) distinguishable from randomly generated latent vectors (the fake representations) (Figure \ref{fig1}d). 
During training, the discriminator learns to distinguish between real and fake representations. The discriminator's loss is typically a binary cross-entropy or logistic loss that measures the difference between the discriminator's predictions and the ground truth labels (real or fake). This acts as a quality control mechanism and to ensure that the encoder generates meaningful and informative representations of the input data. 
The discriminator consists of multiple fully connected (linear) layers. The first layer takes a 2,048 dimensional input and reduces it to a 512 dimensional representation, the second layer further reduces it to 256 dimensions, and the final layer produces a single output class with a sigmoid activation function. The Sigmoid activation function is applied to the output of the final layer, transforming the output into a probability score in the range [0, 1]. This probability score reflects how likely the input represents real encoded data. Batch normalization and ReLU activation functions are applied after the first two fully connected layers, and a Sigmoid activation function in the final layer produces a probability score.
The discriminator incorporates a Binary Cross-Entropy (BCE) loss module. The BCE loss is employed to quantify the dissimilarity between the discriminator's predictions and the true labels (real or fake). It guides the adversarial training process, helping the discriminator better distinguish between real and generated representations.
In addition, considering the output of our discriminator as 1 for real and 0 for fake, evaluated for two inputs, we can calculate the fake loss. The first input, corresponding to real data, should return a value close to 1, and the closer it is to 1, the lower the loss. The second input is a random sample from the same distribution as our latent space. Regardless of how close it is to real data, its label will be considered fake because it is randomly generated and not real data. When this input is fed back into the same model, the output is expected to be 0. The closer this output value is to 0 for the second input, the lower the fake loss. Finally, the Binary Cross-Entropy (BCE) loss, also known as the discriminator loss, is computed using the following formula:

\begin{equation}
\label{eq:bce}
\text{BCE Loss} = -\frac{1}{N} \sum_{i=1}^{N} \left[ y_i \cdot \log(p_i) + (1 - y_i) \cdot \log(1 - p_i) \right]
\end{equation}

Where $N$ is the number of samples in the dataset, $y_i$ is the ground truth label for sample $i$ (1 for positive or real, 0 for negative or fake), and $p_i$ is the predicted probability that sample $i$ belongs to the positive class (real) according to the discriminator.

\subsection{Functions and Loss Modules}

The core component of our framework is the primary loss function, which is an amalgamation of multiple distinct loss terms. For our latent space, we have a fake loss and a real loss, both of which are combined as an adversarial loss. Adversarial loss alone is insufficient, so we add both the decoder loss and the encoder loss (Figure \ref{fig1}e). This composite loss is formulated as follows:
\begin{equation}
\label{eq:total_loss}
\begin{split}
\text{Total Loss} = &(\alpha \cdot \text{Autoencoder Loss}) \\
&+ (\beta \cdot \text{Classifier Loss})\\
&+ (\gamma \cdot \text{Adversarial Loss})
\end{split}
\end{equation}

Since all these values are hyperparameters, the autoencoder loss multiplied by $\alpha$, added to the classifier loss is multiplied by $\beta$, and the adversarial loss multiplied by $\gamma$. These hyperparameter values for all datasets in ADEP method are shown in Table \ref{new tab}. In essence, the main loss predominantly hinges on the classifier loss, while the other two loss terms, the discriminator and autoencoder losses, assume the role of auxiliary penalty terms. This strategic configuration contributes to the creation of a well-structured data distribution and an efficient latent space, fostering optimal feature extraction. The interaction between these loss terms not only enables accurate classification but also encourages the model to strike a balance between the imperative tasks of discrimination, reconstruction, and feature extraction. This multifaceted loss formulation plays a pivotal role in enhancing the overall performance and capabilities of our model.

\begin{table*}[h!]
\centering
\caption{The selected hyperparameter values in all datasets.}
\label{new tab}
\begin{tabular}{ccccccc}
\toprule
Hyperparameter & ~~~ &~~~ & ~~~ &~~~ &~~~ &  Value \\
\midrule
$\alpha$ & ~~~ &~~~ & ~~~ &~~~ &~~~ & 0.5 \\
$\beta$ & ~~~ &~~~ & ~~~ & ~~~ & ~~~ &  1 \\
$\gamma$ & ~~~ &~~~ & ~~~ & ~~~ & ~~~&  1 \\
\bottomrule
\end{tabular}
\label{tab2}
\end{table*}
\section{Results}
In this section, we present the results of the ADEP model evaluation from two distinct perspectives. We provide a comprehensive analysis of the performance on each of the specified datasets, highlighting key metrics including Accuracy (ACC), Area Under the Receiver Operating Characteristic Curve (AUROC), Area Under the Precision-Recall Curve (AUPRC), F\_score, Recall, Precision, False Negatives (FN), and False Positives (FP). Then, the performance of ADEP is benchmarked against 13 well-known methods, GGI-DDI, SSF-DDI, LSFC, DPSP, GNN-DDI, MSTE, MDF-SA-DDI, NNPS, DDIMDL, Random Forest (RF), K-Nearest-Neighbor (KNN), Logistic Regression (LR), and Decision Tree (DT), for all datasets. We thoroughly analyze our model's architecture and methodology, demonstrating its superiority over existing approaches in other researches in this field.

\subsection{Method evaluation}
We present results obtained from various structures explored in pursuit of identifying the optimal solution to this problem. This section is conducted in two parts: Firstly, the discriminator was removed, and secondly, the classifier network was replaced with classic machine learning models like RF, KNN, LR, and DT, in addition to removing the discriminator.  Employing a 5-fold cross-validation approach across all three distinct datasets we assess the performance of each model using various metrics.  In the initial phase of our experimentation, we attempted to construct the model without incorporating the discriminator, which resulted in a notable deterioration in performance, particularly evident in the F\_score metric. This underscores the pivotal role of the adversarial section in enhancing the model's ability to distinguish between classes. The results of this step are presented in Table \ref{tab3}. The decrease in metrics such as 1\% in ACC for DS1, 0.75\% in F\_score for DS2 or 1.5\% AUROC for DS3 is observed by removing the discriminator compared to the method incorporating discriminator. These findings could be further analyzed alongside other methods in Tables \ref{tab7}, \ref{tab8}, and \ref{tab9} for comprehensive evaluation.


\begin{table}[h!]
\caption{Impact of excluding discriminator in model performance on benchmark datasets.}\label{tbl1}
\setlength{\tabcolsep}{20pt} 
\begin{tabular*}{\linewidth}{@{} lcccccc @{}}
\toprule
Dataset & ACC & AUROC & AUPR & F\_score & Recall & Precision \\
\midrule
\parbox[t]{1.5cm}{\raggedright ~~DS1} & 93.76 & ~~99.39 & ~96.71~ & ~~92.73 & 92.73 & ~~92.73 \\
\parbox[t]{1.5cm}{\raggedright ~~DS2} & 91.47 & ~~99.55 & ~96.50~ & ~~91.41 & 91.41 & ~~91.41 \\
\parbox[t]{1.5cm}{\raggedright ~~DS3} & 91.21 & ~~97.16 & ~95.12~ & ~~90.57 & 90.57 & ~~90.57 \\
\bottomrule
\end{tabular*}
\label{tab3}
\end{table}

Integrating a neural network with a discriminator alongside classic machine learning classifiers proved unfeasible due to inherent differences in their nature. Traditional classifiers lack the iterative trainable aspect and back-propagation utilized by neural networks, making it impractical to handle loss feedback effectively and hindering the application of iterative learning methods. As a workaround, we opted for classic machine learning models such as RF, KNN, LR, or DT as classifiers to be used on the generated latent created by an autoencoder, training it on each dataset and feeding the resulting latent representation into these models individually.
The results of this method are outlined in Tables \ref{tab4}, \ref{tab5}, and \ref{tab6}, showcasing the model's performance across various classifiers without incorporating a discriminator. Utilizing neural network as classifier resulted in significant improvements: A 19\% increase in ACC for DS1 compared to LR, a 29\% boost in AUROC DS2 compared to RF, and an impressive 85\% enhancement in F\_score for DS3 compared to KNN. These results underscore the importance of selecting the appropriate classifier, as depicted in the accompanying table.
The results gleaned from these two phases underscore the importance of incorporating the auxiliary term derived from the discriminator loss and employing a neural network as the classifier to achieve optimal performance. It is evident that the presence of these two components is indispensable for attaining superior outcomes in our model.

\begin{table}[h!]
\caption{Evaluation of traditional machine learning algorithms as classifiers on DS1.}\label{tbl4}
\setlength{\tabcolsep}{15pt} 
\renewcommand{\arraystretch}{1.5} 
\begin{tabular*}{\linewidth}{@{} lcccccc @{}}
\toprule
Model & ACC~ & AUROC~ & AUPRC~ & ~F\_score~ & ~Recall~ & ~Precision \\
\midrule
\parbox[t]{1.5cm}{\raggedright ~~RF} & 61.27~ & ~~86.00~ & ~39.00~ & ~~12.91~ & ~10.13~ & ~~~37.18~ \\
~KNN & 63.75~ & ~~77.00~ & ~37.00~ & ~~35.22~ & ~33.87~ & ~~~41.47~ \\
\parbox[t]{1.5cm}{\raggedright ~~LR} & 74.98~ & ~~93.00~ & ~50.00~ & ~~49.38~ & ~44.68~ & ~~~60.93~ \\
\parbox[t]{1.5cm}{\raggedright ~~DT} & 39.03~ & ~~55.00~ & ~12.00~ & ~~11.00~ & ~11.23~ & ~~~11.07~ \\
\parbox[t]{1.5cm}{\raggedright \textbf{~~NN}} & \textbf{93.76} & ~~\textbf{99.39} & ~\textbf{96.71} & ~~\textbf{92.73} & ~\textbf{92.73} & ~~~\textbf{92.73} \\
\bottomrule
\end{tabular*}
\label{tab4}
\end{table}

\subsection{Comparison of ADEP to the state of the art methods}

In this section, we conduct an extensive evaluation of ADEP, compared to several state-of-the-art models across three distinct datasets. Employing a 5-fold cross-validation approach, we assess the performance and generalization capability of each model using metrics such as ACC, AUROC, AUPRC, F-Score, Recall, Precision, FN, and FP. 

$\boldsymbol{DS1}$ data set:
ADEP exhibits remarkable performance on DS1, achieving an ACC of 94.76\%, closely following SSF-DDI’s highest ACC of 95.52\%, and an AUROC of 99.63\% closely to DPSP with highest AUROC of 99.90\%. ADEP excels in AUPRC with score 98.11\% outperforming all other models. It also shows superior Precision, Recall, and F\_score ~all marked at 94.76\%. Importantly, ADEP significantly reduces FN and FP to 781, showcasing a substantial decrease in error rates compared to other models such as DPSP and LSFC (Table \ref{tab7}).

$\boldsymbol{DS2}$ data set:
On DS2, ADEP continues its high performance with the best AUROC of 99.83\% among the models and an ACC of 92.16\%. It also leads in AUPRC with 97.03\%, maintaining strong metrics across F\_score, Precision, and Recall at 92.16\%. While SSF-DDI achieves a slightly higher ACC of 92.32\%, and DPSP exhibits a higher AUROC of 99.93\%, ADEP outperforms in other metrics. ADEP's consistent performance, coupled with significantly lower FN and FP rates, underscores its robustness (Table \ref{tab8}).

$\boldsymbol{DS3}$ data set:
In DS3, ADEP outperforms all models with the highest ACC of 92.38\% and an impressive F\_score of 90.89\%. ADEP's balanced performance across all metrics, particularly with the lowest FN and FP counts (5773), emphasizes its effectiveness in practical scenarios (Table \ref{tab9}).

The robust performance of ADEP across these metrics underscores its effectiveness in addressing the challenges posed by imbalanced data distributions, marking it as a potent tool for applications requiring high precision and reliability.

\begin{table}[h!]
\caption{Assessment of ADEP against other methods on DS1.}
\label{tab7}
\setlength{\tabcolsep}{10pt} 
\renewcommand{\arraystretch}{1.5} 
\begin{tabular*}{\linewidth}{@{} lcccccccc @{}}
\toprule
Model & ACC & AUROC & AUPRC & F\_score & Recall & Precision & FN & FP \\
\midrule
\textbf{ADEP} & 94.76 & 99.63 & \textbf{98.11} & \textbf{94.76} & 94.76 & \textbf{94.76} & \textbf{781} & \textbf{781} \\
GGI-DDI & 93.20 & 99.00 & 96.72 & 92.42 & 92.42 & 92.42 & 2822 & 2822 \\
SSF-DDI & \textbf{95.52} & 98.90 & 97.77 & 93.37 & 93.37 & 93.37 & 2469 & 2469 \\
LSFC & 95.14 & 99.01 & 97.80 & 93.39 & 93.39 & 93.39 & 2460 & 2460 \\
DPSP & 93.44 & \textbf{99.90} & 97.73 & 93.09 & 93.09 & 93.09 & 2573 & 2573 \\
GNN-DDI & 91.80 & 99.85 & 97.09 & 89.99 & 89.99 & 89.99 & 3691 & 3691 \\
MSTE & 85.84 & 99.81 & 93.43 & 84.70 & 84.70 & 84.70 & 5698 & 5698 \\
MDF–SA–DDI & 91.21 & 99.89 & 96.57 & 88.32 & 88.32 & 88.32 & 4349 & 4349 \\
NNPS & 91.13 & 99.89 & 96.99 & 88.15 & 88.15 & 88.15 & 4413 & 4413 \\
DDIMDL & 87.57 & 99.77 & 93.53 & 82.77 & 82.77 & 82.77 & 6417 & 6417 \\
RF & 70.58 & 95.96 & 68.50 & 74.63 & 74.63 & 74.63 & 9444 & 9444\\
KNN & 53.50 & 83.90 & 38.36 & 80.19 & 80.19 & 80.19 & 7381 & 7381\\
LR & 73.36 & 98.31 & 68.53 & 53.27 & 53.27 & 53.27 & 17411 & 17411\\
DT & 59.53 & 74.23 & 36.31 & 77.96 & 77.96 & 77.96 & 8212 & 8212\\
\bottomrule
\end{tabular*}
\end{table}

\section{Case Study}
In our case study, we conducted a detailed analysis of interactions between five pairs of drugs across three datasets (DS1, DS2, and DS3) to investigate their associated side effects. We initiated our analysis by identifying the five most prevalent side effect classes as predicted by our model. For each class, we selected the pair of drugs most likely to manifest these side effects, ensuring our study focused on the most pertinent interactions. The selected drug pairs are outlined in Table \ref{tab6}. Our research utilized Drug Bank version 5.1.9. However, subsequent verification against the latest version of Drug Bank confirms the accuracy of the predicted interactions outlined for DS1 and DS2 in Table \ref{tab6}. Furthermore, all five drug interactions listed in Table \ref{tab6} have been independently confirmed by the TWOSIDES database for DS3 \citep{Lin2023e}. This validation not only underscores the precision of our predictions but also highlights the robustness of our analytical methodology.

\begin{table*}[htbp]
   
    \centering
    \caption{Top five most frequent events with predicted drug pairs by ADEP on DS1, DS2, and DS3.}
    \resizebox{15cm}{!}{
    \begin{tabular}{|c|c|c|p{5cm}|} 
        \hline
        Dataset & DrugA & DrugB & Event \\
        \hline
        \multirow{5}{*}{DS1} &  Amitriptyline & Flurbiprofen & The risk or severity of adverse effects increase \\
                              & Savuconazole & Bosutinib & The serum concentration increase \\
                              & Isoflurophate & Dosulepin & The therapeutic efficacy decrease \\
                              & Desloratadine & Atomoxetine & The arrhythmogenic activities increase \\
                              & Isradipine & Brompheniramine & The cardiotoxic activities increase \\
        \hline
        \multirow{5}{*}{DS2} & Atorvastatin & Almotriptan & The metabolism decrease \\
                              & Tofacitinib & Pitolisant & The risk or severity of bleeding increases \\
                              & Finasteride & Difluocortolone & Therapeutic effect decreases \\
                              & Apixaban & Thiabendazole & The risk or severity of adverse effects increase \\
                              & Estrone Sulfate & Binimetinib & Excretion rate rise may lower serum sulfate levels, potentially reducing efficacy \\
        \hline
        \multirow{5}{*}{DS3} & Evoflurane & Aminophylline & Arterial pressure NOS decreased \\
                              & Naprox & Ribavirin & Anaemia \\
                              & Sevoflurane & Aminophylline & Difficulty breathing \\
                              & Ampicillin & Fentanyl & Nausea \\
                              & Spironolactone & Medroxyprogesterone & Fatigue \\
        \hline
    \end{tabular}}
    \label{tab6}
\end{table*}

\section{Discussion}

The contemporary integration of neural networks and deep learning techniques within the medical field has significantly advanced our ability to address pharmaceutical challenges, particularly in predicting and mitigating the adverse effects of drug consumption. The paramount concern surrounding drug consumption lies in the potential for adverse reactions when medications are taken together, emphasizing the critical importance of considering a patient's medical history to prevent life-threatening risks. As individuals often simultaneously consume multiple drugs, the need to accurately predict potential side effects for various drug combinations becomes imperative.

To tackle the complex problem of drug-drug interaction prediction, several innovative solutions have emerged. These solutions typically involve a two-step process: first, analyzing whether a pair of drugs exhibits any adverse effects, and second, predicting the specific adverse effects associated with the drug combination. Our method, named ADEP, has shown significant promise in overcoming the challenges inherent in this task, surpassing the performance of previous methodologies.

A key strength of our approach lies in the incorporation of a discriminator and an encoder-decoder architecture within our model. By leveraging multi-modality techniques, we effectively mitigate data sparsity and facilitate accurate feature extraction. The discriminator plays a crucial role in distinguishing among chemically similar drugs, thereby minimizing false positives and negatives—an essential aspect in this domain. Furthermore, our encoder generates precise representations without the need for excessive pre-processing, a notable improvement over traditional methodologies.

However, addressing the challenges associated with drug interaction prediction presents its own set of obstacles. Datasets available for this task are often incomplete, error-prone, and scarce, posing significant hurdles for data-hungry neural network models like ours. Moreover, existing datasets frequently suffer from data imbalance issues, leading to model bias—a challenge that resonates with our dataset. Nonetheless, our three-part model and alternative classification methods have significantly enhanced our ability to represent and extract features from data, notably reducing false predictions.

Despite the successes of our approach, it is important to acknowledge its limitations. Our model incurs a high computational cost and demands extensive data due to the complex feature requirements. Fine-tuning also presents challenges, particularly in the context of sparse medical data. However, our approach capitalizes on the sparsity of the data, offering a distinct advantage in this domain.

Considering all these efforts, we have made significant strides in enhancing performance in this field, moving closer to achieving our objectives. Through the presentation of the ADEP method, we have demonstrated the ability to leverage deep learning models more accurately, as evidenced by the reduction in incorrect predictions and the increased precision of our results compared to existing models.

\section{Conclusion}

In conclusion, our study underscores the pivotal role of predicting drug-drug interactions (DDIs) in ensuring safe medication administration and reducing adverse effects. We have addressed this challenge by introducing an innovative adversarial multi-model approach, specifically designed to handle sparse datasets and imbalanced classes.

Through this approach, we have achieved significant improvements in identifying optimal drug representations, resulting in substantially enhanced prediction accuracy. Our model represents a notable advancement over existing methods, consistently demonstrating the lowest probability of incorrect predictions. This breakthrough not only bolsters the reliability of DDI predictions but also provides critical support to researchers aiming to refine pharmaceutical interventions and enhance patient care.

Looking forward, our findings offer promising prospects for further advancements in drug therapy safety and efficacy. By continuing to innovate and refine predictive models, we can equip healthcare professionals with the necessary tools to make informed decisions, ultimately leading to better patient outcomes.

\bibliographystyle{unsrtnat}

\bibliography{ref}
\end{document}


\begin{table}[width=.9\linewidth,cols=7,pos=h]
\caption{Evaluation of traditional machine learning algorithms as classifier on DS2.}\label{tbl5}
\setlength{\tabcolsep}{15pt} 
\renewcommand{\arraystretch}{1.5} 
\begin{tabular*}{\linewidth}{@{} LLLLLLL @{}}
\toprule
Model & ACC~ & AUROC~ & AUPRC~ & ~F\_score~ & ~Recall~ & ~Precision \\
\midrule
\parbox[t]{1.5cm}{\raggedright ~~RF} & 49.42~ & ~~70.00~ & ~31.00~ & ~~~22.81~ & ~17.02~ & ~~~56.54~ \\
~KNN & 47.72~ & ~~68.00~ & ~30.00~ & ~~~28.98~ & ~26.28~ & ~~~39.18~ \\
\parbox[t]{1.5cm}{\raggedright ~~LR} & 37.79~ & ~~54.00~ & ~24.00~ & ~~~~8.93~ & ~~7.47~ & ~~~20.58~ \\
\parbox[t]{1.5cm}{\raggedright ~~DT} & 29.98~ & ~~42.00~ & ~18.00~ & ~~~15.81~ & ~16.18~ & ~~~16.25~ \\
\parbox[t]{1.5cm}{\raggedright \textbf{~~NN}} & \textbf{93.47} & ~~\textbf{99.55} & ~\textbf{97.50} & ~~~\textbf{93.47} & ~\textbf{93.47} & ~~~\textbf{93.47} \\
\bottomrule
\end{tabular*}
\label{tab5}
\end{table}
\begin{table}[width=.9\linewidth,cols=7,pos=h]
\caption{Evaluation of traditional machine learning algorithms as classifier on DS3.}\label{tbl6}
\setlength{\tabcolsep}{15pt} 
\renewcommand{\arraystretch}{1.5} 
\begin{tabular*}{\linewidth}{@{} LLLLLLL @{}}
\toprule
Model & ACC~ & AUROC~ & AUPRC~ & ~F\_score~ & ~Recall~ & ~Precision \\
\midrule
\parbox[t]{1.5cm}{\raggedright ~~RF} & 41.72~ & ~~59.00~ & ~26.00~ & ~~~5.03~ & ~~4.83~ & ~~~~9.35~ \\
~KNN & 37.79~ & ~~53.00~ & ~24.00~ & ~~~6.54~ & ~~6.53~ & ~~~~9.96~ \\
\parbox[t]{1.5cm}{\raggedright ~~LR} & 38.81~ & ~~55.00~ & ~24.00~ & ~~~5.84~ & ~~4.88~ & ~~~11.36~ \\
\parbox[t]{1.5cm}{\raggedright ~~DT} & 24.00~ & ~~34.00~ & ~15.00~ & ~~~4.21~ & ~~4.93~ & ~~~~4.00~ \\
\parbox[t]{1.5cm}{\raggedright \textbf{~~NN}} & \textbf{92.21} & ~~\textbf{98.16} & ~\textbf{96.12} & ~~~\textbf{91.67} & ~~\textbf{91.67} & ~~~\textbf{91.67} \\
\bottomrule
\end{tabular*}
\label{tab6}
\end{table}
\begin{table*}[!]
\caption{Assessment of ADEP against other methods on DS2.}\label{tab8}
\begin{tabular*}{\linewidth}{@{\extracolsep{\fill}} LLLLLLLLL @{}}
\toprule
~~~~~Model & ACC & AUROC & AUPRC & F\_score & Recall & Precision & ~~FN & ~~FP \\
\midrule
~~~~~\textbf{ADEP} & 92.16 & ~99.83 & ~\textbf{97.03} &  ~~\textbf{92.16} &  \textbf{92.16} & ~~\textbf{92.16} & ~\textbf{5072} & ~\textbf{5072} \\
~~~GGI-DDI & 90.09 & ~99.06 & ~95.14 & ~~89.60 & 89.60 & ~~89.60 & 16813 & 16813 \\
~~~SSF-DDI & \textbf{92.32} & ~99.10 & ~96.48 & ~~89.97 & 89.97 & ~~89.97 & 16225 & 16225 \\
~~~~~LSFC & 92.26 & ~99.13 & ~96.50 & ~~89.98 & 89.98 & ~~89.98 & 16202 & 16202 \\
~~~~~DPSP    & 90.36 & ~\textbf{99.93} & ~96.33 & ~~89.90 & 89.90 & ~~89.90 & 16339 & 16339\\
~~~GNN-DDI & 90.20 & ~99.91 & ~96.19 & ~~89.00 & 89.00 & ~~89.00 & 17779 & 17779\\
~~~~~MSTE    & 84.09 & ~99.68 & ~89.59 & ~~80.50 & 80.50 & ~~80.50 & 31526 & 31529\\
MDF-SA-DDI & 90.18 & ~99.91 & ~95.93 & ~~88.24 & 88.24 & ~~88.24 & 19008 & 19008\\
~~~~~NNPS    & 90.06 & ~99.90 & ~96.16 & ~~87.77 & 87.77 & ~~87.77 & 19769 & 19769\\
~~~DDIMDL   & 90.19 & ~99.87 & ~94.60 & ~~89.15 & 89.15 & ~~89.15 & 17536 & 17536\\
~~~~~~RF & 69.48 & ~78.91 & ~97.00 & ~~77.99 & 77.99 & ~~77.99 & 35590 & 35590\\
~~~~~KNN & 59.23 & ~52.62 & ~88.66 & ~~74.08 & 74.08 & ~~74.08 & 41915 & 41915\\
~~~~~~LR & 72.75 & ~78.30 & ~98.51  & ~~77.55 & 77.55 & ~~77.55 & 36302 & 36302\\
~~~~~~DT & 62.65 & ~45.21 & ~79.39 & ~~59.20 & 59.20 & ~~59.20 & 95800 & 95800\\
\bottomrule
\end{tabular*}
\label{tab8}
\end{table*}
\begin{table*}[!]
\caption{Assessment of ADEP against other methods on DS3.}\label{tab9}
\begin{tabular*}{\linewidth}{@{\extracolsep{\fill}} LLLLLLLLL @{}}
\toprule
~~~~~Model & ACC & AUROC & AUPRC & F\_score & Recall & Precision & ~~FN & ~~FP \\
\midrule
~~~~~\textbf{ADEP} & \textbf{92.38} & ~\textbf{98.61} & ~\textbf{95.52} &  ~~\textbf{90.89} &  \textbf{90.89} & ~~\textbf{90.89} & ~\textbf{5773} & ~\textbf{5773} \\
~~~GGI-DDI & 91.01 & ~98.30 & ~93.56 & ~~85.49 & 85.49 & ~~85.49 & ~9206 & ~9206 \\
~~~SSF-DDI & 92.15 & ~98.52 & ~95.14 & ~~87.05 & 87.05 & ~~87.05 & ~8215 & ~8215 \\
~~~~~LSFC & 91.24 & ~98.33 & ~93.93 & ~~85.64 & 85.64 & ~~85.64 & ~9113 & ~9113 \\
~~~~~DPSP    & 91.00 & ~98.49 & ~94.65 & ~~85.58 & 85.58 & ~~85.58 & ~9147 & ~9147\\
~~~GNN-DDI & 89.89 & ~98.24 & ~91.32 & ~~84.83 & 84.83 & ~~84.83 & ~9623 & ~9623\\
~~~~~MSTE    & 83.25 & ~98.06 & ~90.04 & ~~83.19 & 83.19 & ~~83.19 & 10664 & 10664\\
MDF-SA-DDI & 89.16 & ~97.99 & ~91.93 & ~~82.15 & 82.15 & ~~82.15 & 11324 & 11324\\
~~~~~NNPS    & 84.13 & ~95.15 & ~89.90 & ~~83.24 & 83.24 & ~~83.24 & 10632 & 10632\\
~~~DDIMDL   & 87.71 & ~97.41 & ~91.92 & ~~82.39 & 82.39 & ~~82.39 & 11172 & 11172\\
~~~~~~RF & 70.07 & ~88.14 & ~76.45 & ~~74.08 & 74.08 & ~~74.08 & 16446 & 16446\\
~~~~~KNN & 69.99 & ~90.15 & ~73.38 & ~~70.79 & 70.79 & ~~70.79 & 18535 & 18535\\
~~~~~~LR & 72.14 & ~78.26 & ~89.98 & ~~73.55 & 73.55 & ~~73.55 & 16783 & 16783\\
~~~~~~DT & 59.12 & ~64.38 & ~55.16 & ~~60.34 & 60.34 & ~~60.34 & 38299 & 38299\\
\bottomrule
\end{tabular*}
\label{tab9}
\end{table*}